\documentclass[letterpaper]{article} 
\usepackage{aaai24}  
\usepackage{times}  
\usepackage{helvet}  
\usepackage{courier}  
\usepackage[hyphens]{url}  
\usepackage{graphicx} 
\usepackage{natbib}  
\usepackage{caption} 
\usepackage{amsmath}
\usepackage{multirow}
\usepackage{amssymb}
\usepackage{booktabs}
\usepackage{algorithm}
\usepackage{algorithmic}

\usepackage{newfloat}
\usepackage{listings}
\DeclareCaptionStyle{ruled}{labelfont=normalfont,labelsep=colon,strut=off} 
\floatstyle{ruled}
\newfloat{listing}{tb}{lst}{}
\floatname{listing}{Listing}
\title{TDeLTA: A Light-weight and Robust Table Detection Method based on Learning Text Arrangement}
\author{
    Yang Fan\textsuperscript{\rm 1}
    Xiangping Wu\textsuperscript{\rm 1, \rm 2}\thanks{Corresponding authors}
    Qingcai Chen\textsuperscript{\rm 1, \rm 3}\footnotemark[1]
    Heng Li\textsuperscript{\rm 1}\\
    Yan Huang\textsuperscript{\rm 4}
    Zhixiang Cai\textsuperscript{\rm 4}
    Qitian Wu\textsuperscript{\rm 4}
}
\affiliations{
    \textsuperscript{\rm 1}Harbin Institute of Technology (Shenzhen)\\
    \textsuperscript{\rm 2}The Hong Kong Polytechnic University\\
    \textsuperscript{\rm 3}Peng Cheng Laboratory \\
    \textsuperscript{\rm 4}China Mobile Information Technology Co.,Ltd.

    yfan@stu.hit.edu.cn,
    wxpleduole@gmail.com,
    qingcai.chen@hit.edu.cn,
    hengli.lh@outlook.com \\
}

\usepackage{bibentry}

\begin{document}

\maketitle

\begin{abstract}
The diversity of tables makes table detection a great challenge, leading to existing models becoming more tedious and complex. Despite achieving high performance, they often overfit to the table style in training set, and suffer from significant performance degradation when encountering out-of-distribution tables in other domains. To tackle this problem, we start from the essence of the table, which is a set of text arranged in rows and columns. Based on this, we propose a novel, light-weighted and robust \textbf{T}able \textbf{De}tection method based on \textbf{L}earning \textbf{T}ext \textbf{A}rrangement, namely \textbf{TDeLTA}. TDeLTA takes the text blocks as input, and then models the arrangement of them with  a sequential encoder and an attention module. To locate the tables precisely, we design a text-classification task, classifying the text blocks into 4 categories according to their semantic roles in the tables. Experiments are conducted on both the text blocks parsed from PDF and extracted by open-source OCR tools, respectively. Compared to several state-of-the-art methods, TDeLTA achieves competitive results with only 3.1M model parameters on the large-scale public datasets. Moreover, when faced with the cross-domain data under the 0-shot setting, TDeLTA outperforms baselines by a large margin of nearly 7\%, which shows the strong robustness and transferability of the proposed model.
\end{abstract}

\section{Introduction}
\begin{figure}[t]
    \centering
    \includegraphics[width=8cm]{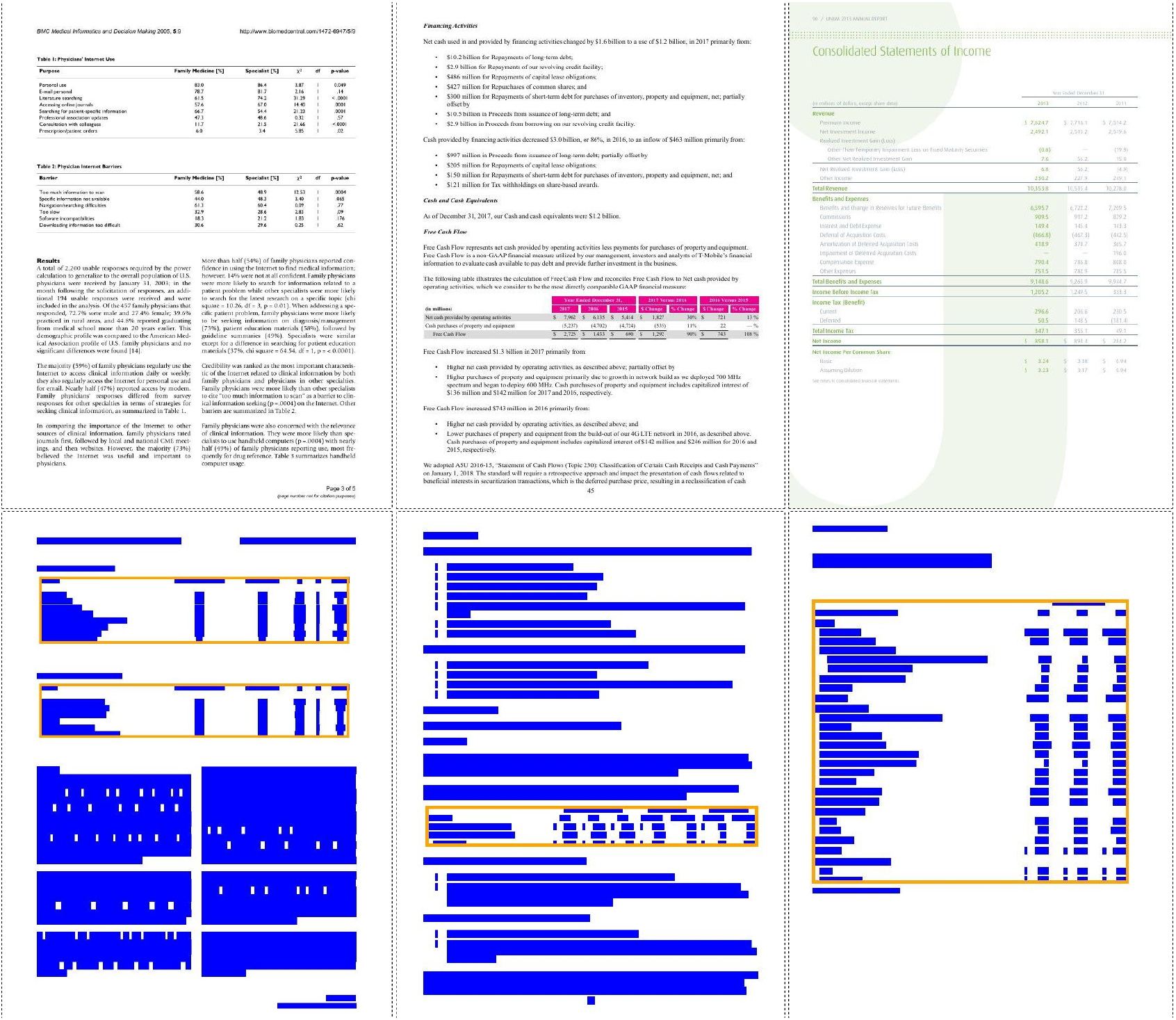}
    \vspace{-2mm}
    \caption{Tables in different styles and their text arrangement. The three images above show three tables in different styles, and the ones below show their text blocks with blue rectangles. The orange boxes denote the location of tables.}
    \vspace{-4mm}
    \label{fig:intro}
\end{figure}
As a popular means of storing and displaying structured data, tables are widely used in various documents including financial reports, scientific articles, invoices, etc. With the rapid growth of the number of digital documents on the Internet, how to teach machines to understand the tables and utilize massive valuable information in them, has become the focus of researchers. Table detection task aims to detect the boundaries of tables in documents. It can be a difficult problem due to the diversity of table styles. For example, some tables have complete borders between the cells of each row and column, while others may have only partial borders, such as the three-line tables in scientific articles. What's more, the diversity of table contents further complicates this issue. 

Research on table detection tasks began in the last century. Early researchers used mostly heuristic-based methods \cite{Gatos2005AutomaticTD,Tupaj1996ExtractingTI,Wang2001AutomaticTG}, which tend to require a lot of manual work and can hardly generalized on tables in different styles. Recently, with the great success of deep learning technology in computer vision applications \cite{Tian_2019_CVPR, DBLP:journals/pami/TianCLJLZYYJ23, zhang2020text, peng2023hierarchical}, many researchers perceive table detection task as a special object-detection problem, taking an document image as input and detecting bounding boxes of tables. 
These methods only treat tables as visual patterns, overlooking the fact that tables are a form of structured data, containing important information in their content organization. As visual models become larger and more complex, this not only increases the inference time and memory usage, limiting the practicality of this technique in many scenarios, such as mobile applications and large-scale data processing, but also makes the model more sensitive to changes in table visual styles.
Although these methods achieved very high performance on some datasets, such as ICDAR 2013 \cite{Gbel2013ICDAR2T}, they often fall into the trap of superficial table styles. 
Experiments show that these image-based methods encounter failure in the scenario of few-shot or 0-shot, similar to other computer vision tasks \cite{tian2020pfenet, tian2022gfsseg, luo2023pfenet++}.


To tackle this problem, we start from the essence of table, a set of text arranged in row and columns. As shown in Figure \ref{fig:intro}, although tables in the upper images have different styles, we can easily find the pattern of them from the text blocks below and then locate them. Motivated by this, we proposed \textbf{TDeLTA}, a light-weighted and robust \textbf{T}able \textbf{De}tection method based on \textbf{L}earning \textbf{T}ext \textbf{A}rrangement, which relies on the position information of text blocks rather than raw images. The fundamental idea of modeling text arrangement enables TDeLTA to be robust to variations in table styles, and demonstrates strong performance when facing out-of-distribution tables from different domains. Additionally, TDeLTA's simple input of text block positions results in a smaller model size, faster inference speed, and reduced memory usage.
We also propose a text-classification task, classifying the text blocks into 4 categories according to their semantic roles, which aids in precise table localization and differentiation between adjacent tables.

Extensive experiments are conducted on two large-scale benchmark datasets (i.e. \emph{PubTables-1M} and \emph{FinTabNet}). It is demonstrated that TDeLTA can achieve competitive performance with only 3.1M parameters compared to the state-of-the-art methods. Moreover, under the 0-shot setting, TDeLTA outperforms these strong baselines by a large margin of nearly 7\%, shows the best robustness and transferability when facing out-of-distribution tables in different domains.

The main contributions of this paper are as follows:
\begin{itemize}
    \item Different from existing image-based methods, we proposed TDeLTA for table detection by learning text arrangement, which takes text blocks as input and can effectively handle out-of-distribution tables in various styles.
    \item To enable precise table localization and help distinguish adjacent tables, we propose a text-classification task, which classifies text blocks into 4 categories according to their semantic roles in tables.
    \item Extensive experiments are conducted on two large-scale benchmark datasets demonstrating the effectiveness and strong robustness of TDeLTA. It is also proven that text arrangement can essentially help the model overcome interference caused by table styles.
\end{itemize}
\begin{figure*}
    \centering
    \includegraphics[width=18cm]{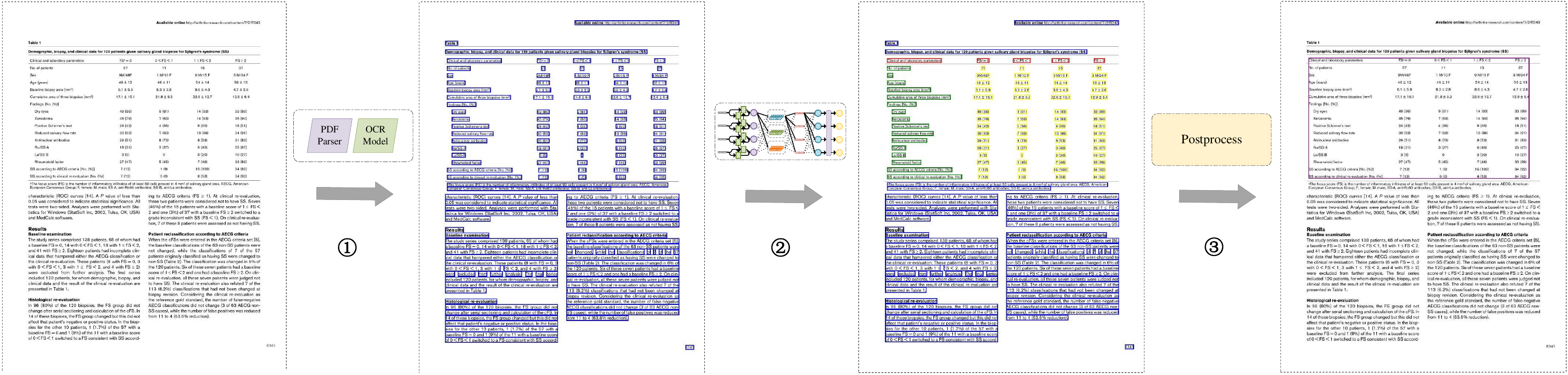}
    \vspace{-2mm}
    \caption{The flowchart of our method. First, we use a PDF parser or a OCR tool to extract text blocks from document pages. Then, we use TDeLTA to classify the text blocks into four categories. At last, we generate the table location with the classification results by a post-processing algorithm}
    \vspace{-4mm}
    \label{fig:flowchat}
\end{figure*}

\section{Related Work}
\subsection{Traditional Methods}
Heuristic-based methods are among the earliest approaches for table detection task. These methods usually employed different visual cues like horizontal and vertical borderlines \cite{Wang2001AutomaticTG,DBLP:journals/ijdar/SeoKC15}, key words \cite{Tupaj1996ExtractingTI}, white space features \cite{Pyreddy1997TINTIAS,DBLP:journals/corr/MacR14}, etc. to detect tables. 
These heuristic-based methods usually require extensive manual efforts to design rules and tune hyper-parameters, and can only work well on documents with uniform layouts. 
\subsection{Deep Learning Based Methods}
With the rapid development of deep learning techniques and the emergence of large-scale table datasets, many deep learning-based methods have been proposed and achieved much better performance, which can be roughly divided into two categories: top-down methods and bottom-up methods.

\paragraph{Top-down methods}
usually treat table detection as a object detection problem and adapt state-of-the-art objection detection frameworks to predict table boundaries. \citeauthor{Hao2016ATD} and \citeauthor{Yi2017CNNBP} used R-CNN based model for table detection, however, the traditional region proposal generation still relied on heuristic rules and handcrafted features. Then, \citeauthor{Vo2018EnsembleOD} and \citeauthor{Gilani2017TableDU} adopted Fast R-CNN and Faster R-CNN to detect tables, and \citeauthor{Gilani2017TableDU} further proposed to use image transformation techniques, such as coloration and dilation, to enhance input documents. \citeauthor{Huang2019AYT} used a Yolo-based method. \citeauthor{Saha2019GraphicalOD} and \citeauthor{Prasad2020CascadeTabNetAA} introduced Mask R-CNN and Cascade Mask R-CNN to table detection task, respectively. \citeauthor{Siddiqui2018DeCNTDD} incorporated deformable convolution and deformable RoI Pooling operations to enhance the robustness of the model when facing geometric transformation problems. \citeauthor{Ma2022RobustTD} proposed CornerNet as a new region proposal network to generate higher quality table proposals for Faster R-CNN.

\paragraph{Bottom-up methods} 
tend to group pixels or page object, such as word and text-line, into table regions. Some researchers view table detection as a semantic segmentation problem. \citeauthor{Yang2017LearningTE} tried to solve this problem in a multi-modal manner. They leveraged both visual features from images and linguistic features from text content to predict a pixel-level segmentation mask. \citeauthor{Kavasidis2018ASC} proposed a saliency-based FCN network performing multi-scale reasoning on visual cues followed by a fully-connected conditional random field (CRF) for localizing tables and charts in digital/digitized documents. Other researchers focused on page objects. \citeauthor{Riba2022TableDI} and \citeauthor{Holecek2019TableUI} both took each document as a graph, where each node represent a text-block and each represent a neighbouring relationship between two nodes. \citeauthor{Holecek2019TableUI} used the coordinates of text-block and the number of characters in each text as the features of nodes. In addition to the geometrical features of the text position, \citeauthor{Riba2022TableDI} also considered the texual features and image features. They both leveraged graph neural networks to encode the text blocks. However, these complex models together with multi-modal features failed to perform well on public datasets.

\section{Methodology}
\subsection{Overview}
Unlike previous works, we transform the table detection task into the classification of text blocks.  As shown in Figure \ref{fig:flowchat}, we first extract the text blocks from the documents via OCR tools or PDF parsers, depending on the type of input files. Then, TDeLTA learns the arrangement among text blocks, and classifies them into four categories according to their semantic roles in tables. Consequently, the boundaries of the tables are generated with the classification results by a post-processing algorithm.
\subsection{Definition of Categories}

As shown in Figure \ref{fig:flowchat}, we divide the text block into 4 categories, including row headers, column headers, content cells, and text outside tables, surrounded by red, green, yellow and blue rectangles respectively. With the help of the categories of row headers and columns, we can easily determine the up and left boundaries of the tables, and thereby divide adjacent tables during the post-processing.

\begin{figure*}
    \centering
    \includegraphics[width=16cm]{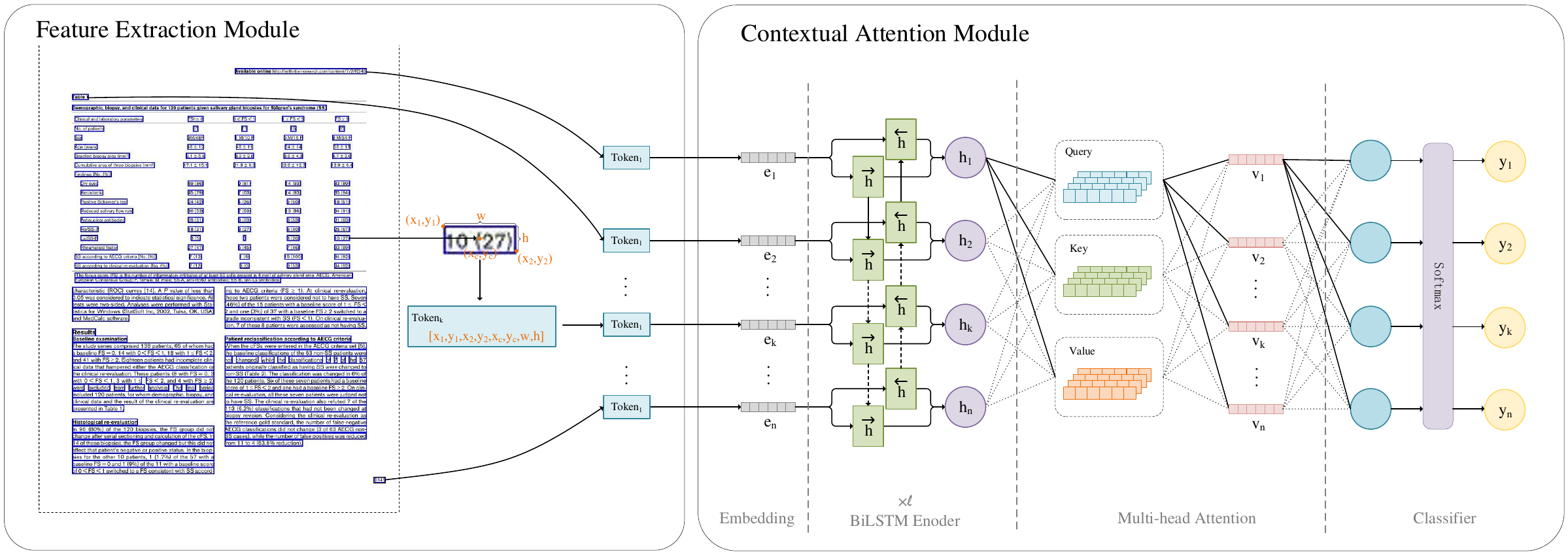}
    \vspace{-2mm}
    \caption{Model Architecture of TDeLTA.}
    \vspace{-4mm}
    \label{fig:model}
\end{figure*}

\subsection{Model Architecture}
As shown in Figure \ref{fig:model}, TDeLTA consists of two modules, namely the Feature Extraction Module and the Contextual Attention Module. 

\subsubsection{Feature Extraction Module}
First, we get the position information of all text blocks from the document pages, represented by $\mathcal{\boldsymbol{D}}=[bbox^{1}, bbox^2, ..., bbox^n]$, where $n$ denotes the number of text blocks and $bbox^{i}$ denotes the coordinates of the $i$-th text blocks. Specifically, $bbox^{i}=[x_1^i, y_1^i, x_2^i, y_2^i]$, where $(x_1^i, y_1^i)$ represents the coordinates of the upper-left corner of the $i$-th text block and $(x_2^i, y_2^i)$ represents the coordinates of the lower-right corner.

Then, we calculate the coordinates of the midpoint, represented by $(x_3^i, y_3^i)$, together with its width $w$ and height $h$, as follows.
\begin{equation}
   \begin{cases}
x_3^i = (x_1^i + x_2^i)/2 \\
y_3^i = (y_1^i + y_2^i)/2 \\
w = x_2^i - x_1^i \\
h = y_2^i - y_1^i
\end{cases}
\label{eq:1}
\end{equation}
After that, we normalize the coordinates as follows.
\begin{equation}
    \begin{cases}
        \hat{x}_k^i = x_k^i / w_D \qquad\qquad k \in \{1, 2, 3\} \\
        \hat{y}_k^i = y_k^i / h_D \qquad\qquad k \in \{1, 2, 3\} \\
        \hat{w} = w / w_D \\
        \hat{h} = h / h_D
    \end{cases}
\end{equation}
where $(w_D, h_D)$ denotes the width and height of the whole document page. Finally, we can get the model input consisting of 8-dimensional features $[\hat{x}_1^i, \hat{y}_1^i,\hat{x}_2^i, \hat{y}_2^i,\hat{x}_3^i, \hat{y}_3^i, \hat{w}, \hat{h}]$ from the coordinates of the $i$-th text block.

\subsubsection{Contextual Attention Module}
The contextual attention module is used to learn the text arrangement, and classify text blocks into 4 pre-defined categories. It has four components, which are the embedding layer, the BiLSTM encoder \cite{Huang2015BidirectionalLM}, the multi-head attention \cite{Vaswani2017AttentionIA} and the linear classifier.

\paragraph{Embedding Layer}
Given an input tensor $\boldsymbol{\rm{I}}\in \mathbb{R}^{n\times8}$, we adopt a linear layer to map it to a high-dimensional embedding space as follows:
\begin{equation}
    \boldsymbol{\rm{E}} = W_e \cdot \boldsymbol{\rm{I}}+b_e
\end{equation}

\paragraph{BiLSTM Encoder} LSTM, as a classic model for sequence encoding, has been widely applied in various tasks. Bidirectional LSTM can learn temporal information from both directions simultaneously. Therefore, we have chosen it as the sequence encoder for TDeLTA. In the subsequent sections, we will provide detailed comparisons with other commonly used encoders.

Give an input $\boldsymbol{\rm{E}}\in \mathbb{R}^{n\times l_I}$, we adopt $N_L$-layer bi-directional long short-term memory network (BiLSTM) to learn the spatial arrangement among text blocks.

\begin{equation}
    \boldsymbol{\rm{H}} = \rm{BiLSTM}(\boldsymbol{\rm{E}})
\end{equation}

Thus, we can get the vector representation $\boldsymbol{\rm{H}} \in \mathbb{R}^{n\times 2l_h}$ that contains spatial arrangement information, where $l_h$ indicates the hidden size of BiLSTM in a single direction. 

\paragraph{Multi-head Attention}
The multi-head attention mechanism has been proven to be effective in mining the connections among tokens in the input sequence on various tasks. Therefore, we use it to further learn the two-dimensional spatial arrangement among text blocks in document pages.

Then, we use the multi-head attention mechanism to further learn the global arrangement information of the text block.

\begin{equation}
    \boldsymbol{\rm{A}} = \rm{MHA}(\boldsymbol{\rm{H}})
\end{equation}
where $\rm{MHA}$ denotes the multi-head attention mechanism. 

\paragraph{Linear Classifier}
Since we get the arrangement-aware vector representation $\boldsymbol{\rm{A}} \in \mathbb{R}^{n\times l_A}$, where $l_A$ denotes the hidden size of the output of multi-head attention layer, a linear classifier layer is used to classify the text blocks as follows:
\begin{equation}
    \hat{Y} = \rm{argmax}(\rm{softmax}(W_{cls}\cdot \boldsymbol{\rm{A}}+ b_{cls}))
\end{equation}
The text blocks in document pages are now classified into four categories.

\paragraph{Loss Function}
As we treat it as a classification task, we choose to use Cross Entropy loss as the loss function of TDeLTA.
\begin{equation}
    \mathcal{L}=-\frac{1}{n}\sum\limits_{i=1}^{n}{y^ilog(\hat{y^i})+(1-y^i)log(1-\hat{y^i})}
\end{equation}

\begin{table*}[ht]
    \centering
    \small
    \begin{tabular}{@{}lccccccccccccc@{}}
    \toprule
    \multirow{2}{*}{Methods} & \multicolumn{3}{c}{IoU@0.5}   & & \multicolumn{3}{c}{IoU@0.6}  & & \multicolumn{3}{c}{Avg. (IoU@0.5-0.95 )} & & \multirow{2}{*}{\#Parm (M)} \\  
    \cline{2-4} \cline{6-8} \cline{10-12}
    &  \multicolumn{1}{c}{P} & \multicolumn{1}{c}{R} & \multicolumn{1}{c}{F1} &&\multicolumn{1}{c}{P} & \multicolumn{1}{c}{R} & \multicolumn{1}{c}{F1} & &\multicolumn{1}{c}{P} & \multicolumn{1}{c}{R} & \multicolumn{1}{c}{F1} & &  \\ 
    \toprule
    \begin{tabular}[c]{@{}c@{}}Table-DETR \citeyearpar{Smock2021PubTables1MTC} \end{tabular} & $\bf99.50$& $\bf99.97$&$\bf99.74$&&$\bf99.50$&$\bf99.96$&$\bf99.73$&&$97.08$&$98.57$&$97.82$& & $28.9$        \\ 
    \begin{tabular}[c]{@{}c@{}}CascadeTabNet \citeyearpar{Prasad2020CascadeTabNetAA} \end{tabular}     &$99.00$&$99.92$&$99.46$&&$99.00$&$99.91$&$99.45$&&$\bf98.99$&$\bf99.83$&$\bf99.41$& &$82.6$                             \\ 
    \begin{tabular}[c]{@{}c@{}}DiT \citeyearpar{Li2022DiTSP} \end{tabular}  &$98.99$&$99.93$&$99.47$&&$98.99$&$99.92$&$99.46$&&$98.96$&$99.81$&$99.38$& &$113.2$         \\ 
    \begin{tabular}[c]{@{}c@{}}YoLov7 \citeyearpar{Wang2022YOLOv7TB}\end{tabular}    &$99.00$&$99.90$&$99.45$&&$99.00$&$99.88$&$99.44$&&$98.77$&$99.52$&$99.14$& & $36.9$ \\ 
    \midrule
    \begin{tabular}[c]{@{}c@{}} TDeLTA \end{tabular}        &$98.53$&$99.71$&$99.12$&&$98.47$&$99.65$&$99.05$&&$98.02$&$99.39$&$98.70$& &$3.1$  \\
    \begin{tabular}[c]{@{}c@{}} \quad w/ Transformer\end{tabular}       &$98.49$&$99.53$&$99.01$&&$98.41$&$99.45$&$98.92$&&$97.80$&$99.11$&$98.45$&&$4.7$  \\
    \begin{tabular}[c]{@{}c@{}} \quad w/ LSTM\end{tabular}       &$97.84$&$99.13$&$98.48$&&$96.67$&$98.92$&$97.78$&&$94.62$&$97.66$&$96.11$& &$4.2$  \\
    \begin{tabular}[c]{@{}c@{}} \quad w/ BiGRU\end{tabular}       &$98.26$&$99.39$&$98.82$&&$98.18$&$99.32$&$98.75$&&$96.49$&$98.46$&$97.47$& &$2.4$ \\
 \bottomrule
    \end{tabular}
    \vspace{-2mm}
    \caption{Performance comparison on PubTables-1M. We report results of TDeLTA with different encoders. }
    \vspace{-4mm}
    \label{tab:ptb}
\end{table*}

\subsection{Post-processing Algorithm}
After getting the classification results of text blocks, we employ a post-processing algorithm to generate table boundaries. The post-processing algorithm consists of two steps: 1) aggregating the neighboring text classified as being inside tables (including three categories) to form preliminary results., and 2) generating the top and left boundaries of the tables based on row headers and column headers, and then splitting the preliminary results to generate the final detected boundaries. We designed the second step to handle cases where multiple tables are adjacent.

\section{Experiments}
\subsection{Datasets}
\paragraph{PubTables-1M\footnote{https://github.com/microsoft/table-transformer}}\citeyearpar{Smock2021PubTables1MTC} contains 460,589 fully-annotated pages for training; 57,591 for validation; and 57,125 for testing. This dataset focuses on tables in scientific domain, as they choose the PMCOA corpus to be their data source, which consists of millions of publicly available scientific articles. The authors generated the labels automatically by matching the text content from PDF documents and XML documents. 
\paragraph{FinTabNet\footnote{https://developer.ibm.com/exchanges/data/all/fintabnet/}}\citeyearpar{Zheng2020GlobalTE} contains 48,001 fully-annotated pages for training; 5,943 for validation; and 5,903 for testing. This dataset focuses on complex tables from the annual reports of the S\&P 500 companies. Financial tables often have very different styles compared with those in scientific documents, like fewer borderlines and more diverse background colors. Therefore, we report results on the test set of FinTabNet under 0-shot setting to verify the robustness and transfer ability of TDeLTA when facing out-of-distribution tables. 

\subsection{Metrics}
Like objection detection tasks, we take Intersection-over-Union(IoU) as our evaluation metrics, where a detection bounding box is considered as a positive result when the proportion of its intersection area with ground truth is larger than the threshold. All results in this experiment were calculated by the open-source Python package Pycocotools.

\subsection{Baselines}
We compare TDeLTA with several strong baselines including Table-DETR~\cite{Smock2021PubTables1MTC}, CascadeTabNet~\cite{Prasad2020CascadeTabNetAA}, DiT~\cite{Li2022DiTSP} and YoLov7~\cite{Wang2022YOLOv7TB}. Table-DETR is a variant of DETR~\cite{carion2020end} on table-related tasks, and the best baseline reported in the original paper of Pubtables-1M. Here we use its officially published weights. CascadeTabNet is a widely-used method based on Cascade R-CNN~\cite{cai2018cascade} and have achieved great performance in many table detection benchmarks. YoLov7 is one of the state-of-the-arts in object detection task. We trained these two baseline models on Pubtables-1M by ourselves. DiT, LayoutLMv3 and StrucTexTv2 are all document-oriented visual backbone pretrained on large-scale unlabeled document images. We fine-tuned DiT on PubTables-1M and utilized the officially released weights fine-tuned on PubLayNet for the other two models, since both datasets derive from the PubMed corpus, we assume they have similar data distributions. Therefore, for LayoutLMv3 and StrucTexTv2, we only conduct comparisons under the 0-shot setting on the FinTabNet dataset.

\subsection{Implementation Details}
For BiLSTM layers, we set the hidden size $l_h$ to 128 and the number of layers $N_L$ to 8. For the multi-head attention layer, we set the number of heads $N_h$ to 4 and the output size $l_A$ to 128 which equals to $l_h$. Ablation study of hyperparameters can be found in the following section. 

For each sequence encoder compared, we set the hidden size to 128 and the number of layers to 8, except for LSTM, which has a hidden size of 256. This ensures that the size of each model is similar for a fair comparison.

All the weights are initialized with a uniform distribution. During training, we employ an Adam~\cite{Kingma2014AdamAM} optimizer for $300$ epochs using a cosine decay scheduler with $10\%$ warmup steps. A batch size of $256$ and an initial learning rate of $0.001$ are used. We implement our approach based on PyTorch and conduct experiments on Nvidia Tesla v100 32G GPUs.

\subsection{Experimental Results}
\subsubsection{Results on PubTables-1M}
As shown in Table~\ref{tab:ptb}, TDeLTA and all the strong baselines achieve very high performance on Pubtables-1M, with the average F1 scores around $99\%$. This is because the documents in Pubtables-1M come from a single source, and the table styles are relatively uniform, making it simpler for these strong baselines. TDeLTA achieves competitive results while the number of parameters is just a fraction of other methods. 

We also compare the BiLSTM encoder with other sequential encoders. (\romannumeral1) LSTM exhibits the poorest performance among the encoders, primarily due to its unidirectional nature. (\romannumeral2) BiGRU achieves a slight lower performance with just 77\% of the parameters (2.4M/3.1M). (\romannumeral3) Transformer relies solely on positional embedding to learn sequential information. BiLSTM, by requiring tokens to be inputted in a specific order, exhibits heightened sensitivity to sequential features. Furthermore, we have incorporated a multi-head attention module in TDeLTA, allowing other sequential encoders to harness the advantages of Transformer.

\begin{table*}[ht]
    \centering
    \small
    \begin{tabular}{@{}lccccccccccccc@{}}
    \toprule
    \multirow{2}{*}{Methods} & \multicolumn{3}{c}{IoU@0.5}   & & \multicolumn{3}{c}{IoU@0.6}  & & \multicolumn{3}{c}{Avg. (IoU@0.5-0.95)} & & \multirow{2}{*}{\#Parm (M)} \\  
    \cline{2-4} \cline{6-8} \cline{10-12}
    &  \multicolumn{1}{c}{P} & \multicolumn{1}{c}{R} & \multicolumn{1}{c}{F1} &&\multicolumn{1}{c}{P} & \multicolumn{1}{c}{R} & \multicolumn{1}{c}{F1} & &\multicolumn{1}{c}{P} & \multicolumn{1}{c}{R} & \multicolumn{1}{c}{F1} & &  \\ 
    \toprule
    \begin{tabular}[c]{@{}c@{}}Table-DETR \citeyearpar{Smock2021PubTables1MTC} \end{tabular} & $53.20$& $79.43$&$63.72$&&$48.75$&$72.60$&$58.33$&&$38.81$&$59.46$&$46.96$& & $28.9$        \\ 
    \begin{tabular}[c]{@{}c@{}}CascadeTabNet \citeyearpar{Prasad2020CascadeTabNetAA} \end{tabular}     &$47.05$&$51.72$&$49.27$&&$44.15$&$49.78$&$46.80$&&$37.23$&$43.37$&$40.06$& &$82.6$                             \\ 
    \begin{tabular}[c]{@{}c@{}}DiT \citeyearpar{Li2022DiTSP} \end{tabular}  &$81.77$&$89.61$&$85.51$&&$78.62$&$87.01$&$82.60$&&$69.11$&$78.37$&$73.45$& &$113.2$         \\ 
    \begin{tabular}[c]{@{}c@{}}YoLov7 \citeyearpar{Wang2022YOLOv7TB}\end{tabular}    &$64.00$&$80.59$&$71.34$&&$62.62$&$79.66$&$70.12$&&$53.14$&$69.96$&$60.40$& & $36.9$ \\ 
    \begin{tabular}[c]{@{}c@{}}LayoutLMv3* \citeyearpar{Huang2022LayoutLMv3PF} \end{tabular}  &$78.49$&$88.82$&$83.34$&&$76.22$&$87.27$&$81.37$&&$65.02$&$77.90$&$70.88$& &$133.0$         \\ 
    \begin{tabular}[c]{@{}c@{}}StrucTexTv2* \citeyearpar{Yu2023StrucTexTv2MV}\end{tabular}    &$64.81$&$92.19$&$76.11$&&$60.70$&$89.23$&$72.25$&&$43.90$&$68.82$&$53.61$& & $50.2$ \\ 
    \midrule
    \multicolumn{5}{l}{\textbf{PDF Parser}} &&&&&&&&& \\
    \midrule
    \begin{tabular}[c]{@{}c@{}} TDeLTA \end{tabular}        &$\bf82.82$&$\bf92.00$&$\bf87.17$&&$\bf80.86$&$\bf90.61$&$\bf85.46$&&$\bf74.88$&$\bf86.93$&$\bf80.46$& &$3.1$  \\
    \begin{tabular}[c]{@{}c@{}} \quad w/ Transformer\end{tabular}       &$74.32$&$86.42$&$79.92$&&$71.47$&$84.86$&$77.59$&&$65.07$&$80.55$&$71.99$& &$4.7$  \\
    \begin{tabular}[c]{@{}c@{}} \quad w/ LSTM\end{tabular}       &$66.29$&$81.74$&$73.21$&&$61.73$&$78.87$&$69.25$&&$50.27$&$70.25$&$58.60$& & $4.2$  \\
    \begin{tabular}[c]{@{}c@{}} \quad w/ BiGRU\end{tabular}       &$78.42$&$88.71$&$83.25$&&$75.08$&$86.85$&$80.54$&&$64.26$&$79.77$&$71.18$&&$2.4$  \\

    \midrule
    \multicolumn{5}{l}{\textbf{Open-source OCR Tool}} &&&&&&&&& \\
    \midrule
    \begin{tabular}[c]{@{}c@{}} TDeLTA \end{tabular}        &$\bf83.46$&$\bf91.60$&$\bf87.34$&&$\bf81.71$&$\bf90.52$&$\bf85.89$&&$\bf72.72$&$\bf84.81$&$\bf78.30$& &$3.1$  \\
    \begin{tabular}[c]{@{}c@{}} \quad w/ Transformer\end{tabular}       &$76.43$&$87.66$&$81.66$&&$73.47$&$85.93$&$79.21$&&$63.58$&$79.09$&$70.49$& & $4.7$  \\
    \begin{tabular}[c]{@{}c@{}} \quad w/ LSTM\end{tabular}       &$74.67$&$85.64$&$79.78$&&$69.83$&$82.76$&$75.75$&&$55.78$&$72.63$&$63.10$&&$4.2$  \\
    \begin{tabular}[c]{@{}c@{}} \quad w/ BiGRU\end{tabular}       &$80.08$&$89.24$&$84.40$&&$75.54$&$86.99$&$80.86$&&$62.40$&$77.76$&$69.24$&&$2.4$  \\
 \bottomrule
    \end{tabular}
    \vspace{-2mm}
    \caption{Performance comparison on FinTabNet under 0-shot setting. We report results of TDeLTA that applying different techniques to extract text blocks (i.e. PDF parsers and open-source OCR tools). * indicates fine-tuned on PubLayNet.}
    \vspace{-4mm}
    \label{tab:FTN}
\end{table*}

\subsubsection{Results on FinTabNet}
To better verify the robustness of TDeLTA on cross-domain data, we directly test these models on the test set of FinTabNet without fine-tuning. Results are listed in Table~\ref{tab:FTN}. Among all the baselines, it is reasonable that DiT and LayoutLMv3 show the best robustness since they have the most parameters and are pre-trained on large-scale unlabeled document images. Although CascadeTabNet achieves the best result on PubTables-1M, its performance decreased severely, and Table-DETR and YoLov7 have also exhibited similar trends. The decline in performance demonstrates that these models excessively focus on the style of tables and overfit to the data in the training set, resulting in poor performance when faced with out-of-distribution data. In contrast, TDeLTA achieves the best performance with the fewest parameters and without using any external data, and the gap becomes larger as the IoU threshold increased, with an average F1 score surpassing DiT by more than $7\%$. This demonstrates that TDeLTA can effectively learn the intrinsic features of tables and exhibits robustness across different table styles.

Considering that in many cases only document images can be obtained without the annotations of text blocks, to demonstrate the practicality of TDeLTA, we utilize the light-weight open-source OCR tool PaddleOCR\footnote{https://github.com/PaddlePaddle/PaddleOCR} to perform text detection on the images from FinTabNet, thereby getting model input from raw images. As we can see in Table~\ref{tab:FTN}, under this setting, TDeLTA's performance only experienced a minor decrease, while still significantly outperforming baseline models. This demonstrates the robustness of TDeLTA, as it does not rely on high-quality text position annotations and can be effectively applied in various scenarios. At lower IoU threshold values, it can be observed that TDeLTA performs better when using OCR results as input slightly. This could be attributed to that some trivial text was missed by the OCR tool, which may introduce interference to the model.

\subsection{Ablation Study}

To further study the influence of model scale on the performance of TDeLTA, we conduct ablation studies on the number of layers and the hidden size of the BiLSTM encoder.

\begin{table}[ht]
    \centering
    \small
    \begin{tabular}{@{}lcccccc@{}}
    \toprule
    \quad\multirow{2}{*}{$N_L$} & \multicolumn{2}{c}{PubTables-1M}   & & \multicolumn{2}{c}{FinTabNet}  &   \multirow{2}{*}{\#Parm (M)} \\  
    \cline{2-3} \cline{5-6}
    &  \multicolumn{1}{c}{AP} & \multicolumn{1}{c}{AR} &&\multicolumn{1}{c}{AP} & \multicolumn{1}{c}{AR}  &  \\ 
    \toprule
    \multicolumn{1}{c}{$2$} & $97.23$& $98.88$&&$67.32$&$81.94$&$0.7$    \\ 
    \multicolumn{1}{c}{$4$} &$97.40$&$98.97$&&$70.91$&$84.07$&$1.5$      \\ 
    \multicolumn{1}{c}{$6$} &$97.69$&$99.17$&&$72.49$&$84.54$&$2.3$         \\ 
    \multicolumn{1}{c}{$8$} &$\bf98.02$&$\bf99.39$&&$74.89$&$86.93$&$3.1$         \\ 
    \multicolumn{1}{c}{$12$} &$97.91$&$99.20$&&$\bf76.08$&$\bf87.10$&$4.7$         \\ 
 \bottomrule
    \end{tabular}
    \vspace{-2mm}
    \caption{Ablation study on the number of layers of BiLSTM }
    \vspace{-4mm}
    \label{tab:ab1}
\end{table}

\paragraph{Ablation study on the number of layers.}
We first keep the hidden size of BiLSTM as 128 and set the number of BiLSTM layers to 2, 4, 6, 8, 12, separately. Results are listed in Table \ref{tab:ab1}. We can observe that as the number of layers increases, the performance of TDeLTA gradually improves on both datasets, and it stabilizes when approaching 8 layers. However, when the number of layers further increases to 12, the performance on the PubTables-1M slightly decreases. This indicates that increasing the number of layers can enhance TDeLTA's fitting ability and reach saturation around 8 layers. Furthermore, it can be seen that when the number of layers is 4 and the number of parameters is 1.5M, the performance on FinTabNet already surpasses all baseline models. Considering the model's performance and efficiency, we choose to set the number of layers to 8.

\begin{table}[ht]
    \centering
    \small
    \begin{tabular}{@{}lcccccc@{}}
    \toprule
    \quad\multirow{2}{*}{$l_h$} & \multicolumn{2}{c}{PubTables-1M}   & & \multicolumn{2}{c}{FinTabNet}  &   \multirow{2}{*}{\#Parm (M)} \\  
    \cline{2-3} \cline{5-6}
    &  \multicolumn{1}{c}{AP} & \multicolumn{1}{c}{AR} &&\multicolumn{1}{c}{AP} & \multicolumn{1}{c}{AR}  &  \\ 
    \toprule
    \multicolumn{1}{c}{$16$} & $88.54$& $94.36$&&$41.66$&$62.69$&$0.051$    \\ 
    \multicolumn{1}{c}{$32$} &$96.08$&$98.37$&&$71.81$&$84.18$&$0.199$      \\ 
    \multicolumn{1}{c}{$64$} &$97.87$&$99.21$&&$73.50$&$85.70$&$0.783$         \\ 
    \multicolumn{1}{c}{$128$} &$\bf98.02$&$\bf99.39$&&$\bf74.89$&$\bf86.93$&$3.1$         \\ 
 \bottomrule
    \end{tabular}
    \vspace{-2mm}
    \caption{Ablation study on the hidden size of BiLSTM}
    \vspace{-4mm}
    \label{tab:ab2}

\end{table}

\paragraph{Ablation study on the hidden size.} 
We fixed the number of layers to 8 and set the hidden size to 16, 32, 64, 128 respectively to conduct ablation experiments. Results are listed in Table \ref{tab:ab2}. We can see that as the hidden size increases, the performance of TDeLTA gradually improves on both datasets. When the hidden size approaches 128, the performance improvement becomes more gradual. From Table \ref{tab:ab2}, it can be observed that when the hidden size is set to 32, the model with only 119k parameters already outperforms all basline models on FinTabNet. This proves the strong transferability and stability of TDeLTA. Considering the model's performance and efficiency, we choose to set the hidden size to 128.

\begin{figure*}
    \centering
    \includegraphics[width=18cm]{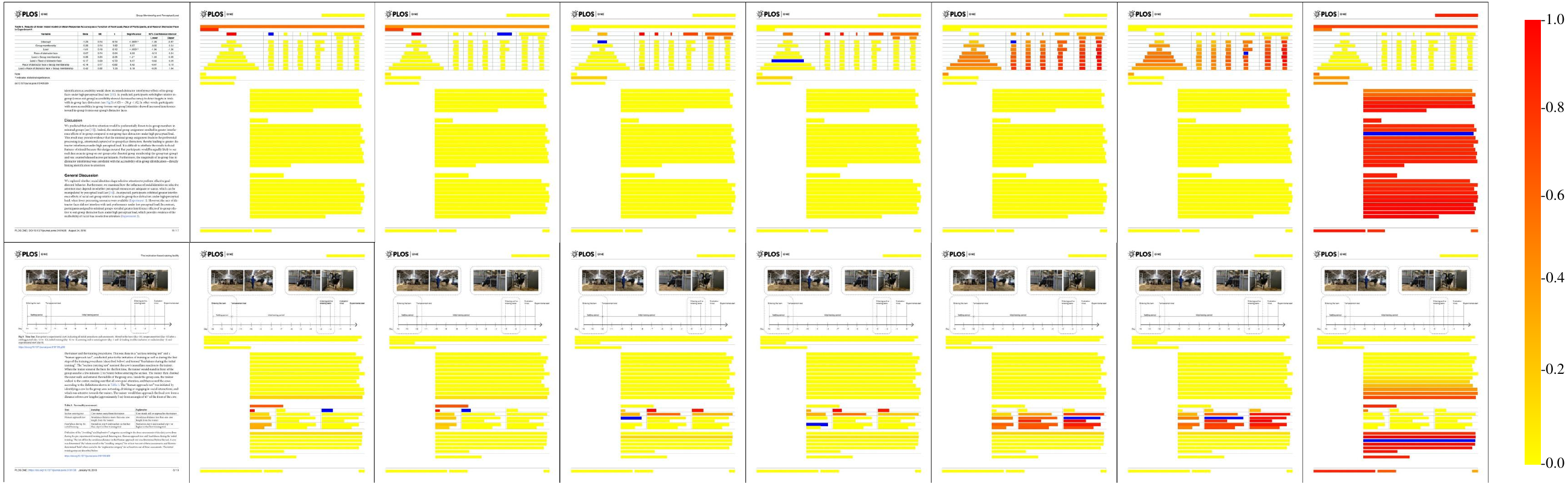}
    \vspace{-2mm}
    \caption{Visualization of attention scores. The first column represents the original inputs. The text block in blue denotes the current step, and each subfigure displays the attention scores of all other text blocks with respect to the blue one.}
    \vspace{-4mm}
    \label{fig:att}
\end{figure*}

\subsection{Visualization of Attention}
For a deeper look, we visualized the attention scores for two sampled documents. As shown in Figure \ref{fig:att}, the first column represents the original inputs. The text block in blue denotes the current step, and each subfigure displays the attention scores of all other text blocks with respect to the blue one, indicating the level of attention the blue text block pays to others in the document. As we can see, different categories of text blocks focus on global information differently. The row headers is the forefront part of the table, which pays more attention to the delineation area; The column headers determines themselves based on the information from row headers; content cells are more concerned with neighboring text arranged in rows and columns; text outside the tables are of little concern for the information within the table. This demonstrates that TDeLTA has learned the semantic roles of each text within the table and their interdependencies, and can utilize them to obtain representations for different texts, indicating that TDeLTA truely understands the tables.

\subsection{Case Study}
\begin{figure}[t]
    \centering
    \includegraphics[width=8cm]{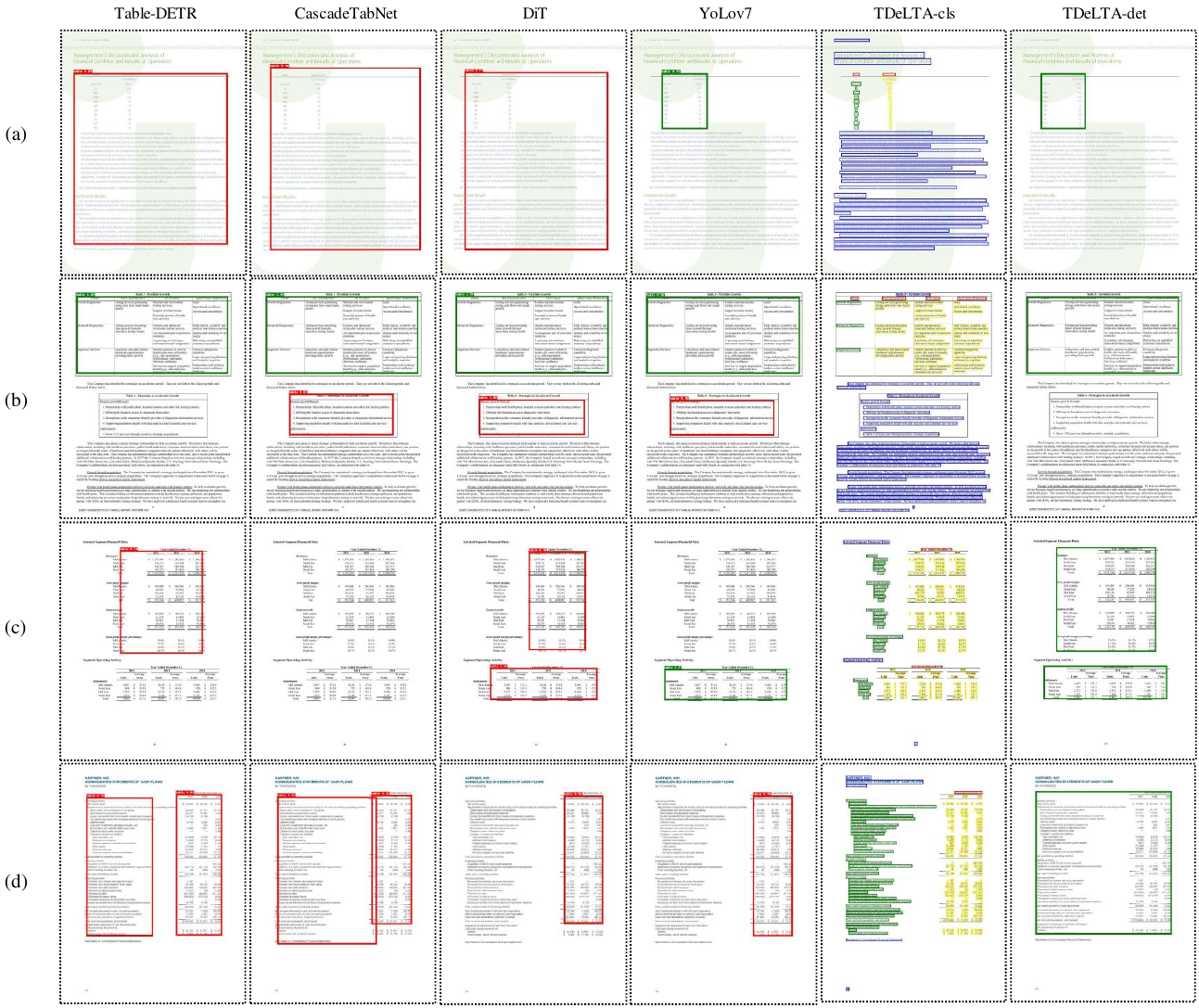}
    \vspace{-2mm}
    \caption{Experimental results on FinTabNet under 0-shot setting. Columns 1-4 show the results of baseline models, respectively. Columns 5-6 represent TDeLTA's classification and detection results. 
    The red boxes represent the bad cases, while the green ones denote correct detection.}
    \vspace{-4mm}
    \label{fig:samples}
\end{figure}

We sample three documents with tables in different styles from the test set of FinTabNet and the results are shown in Figure \ref{fig:samples}. Each row in the figure represents a sampled document. Columns 1-4 show the results of baseline models, respectively. Columns 5-6 represent TDeLTA's classification and detection results. The detection boxes in red represent the bad cases, while those in green are correct results.

Figure \ref{fig:samples} (a) shows that the background pattern of the image can seriously affect the detection results of the image-based baselines. And the horizontal line above the table also misleads them. In Figure \ref{fig:samples} (b), there are some texts surrounded by a rectangular box, and three of these baselines incorrectly detect it as a table, which shows that they learned to detect rectangles, not tables. Figure \ref{fig:samples} (c) illustrates the poor performance of these baseline models when faced with borderless tables. Figure \ref{fig:samples} (d) shows that when the columns in borderless tables is spaced far apart, they are easily recognized as multiple tables by baselines. All these cases emphasizes that these image-based methods do not truly understand tables but rather overfit to shallow visual features. The results in column 5-6 show that TDeLTA can be  robust to variations in table styles and detect correctly.

\subsection{Error Analysis}
\begin{figure}[t]
    \centering
    \includegraphics[width=8cm]{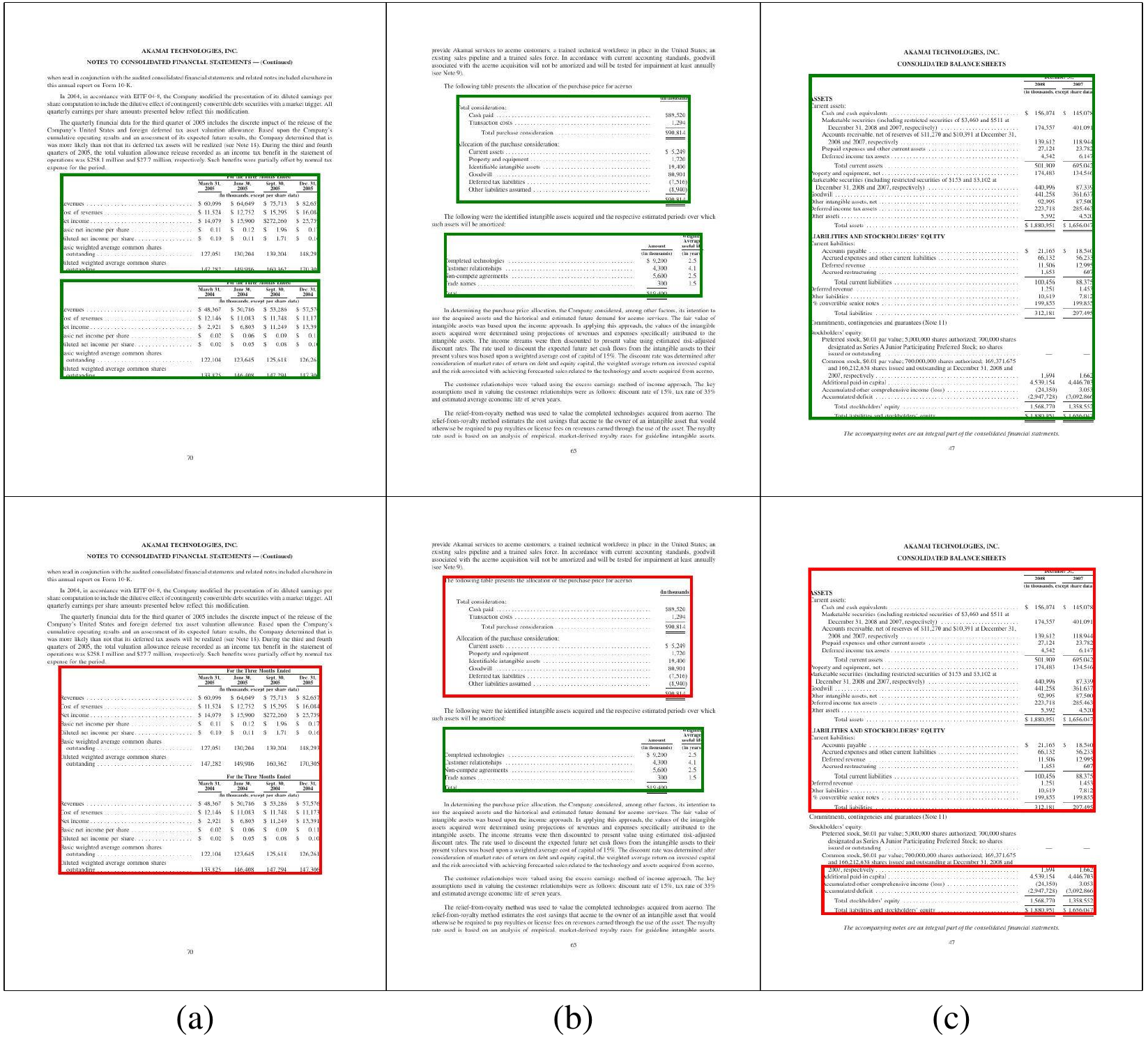}
    \vspace{-2mm}
    \caption{Bad cases of TDeLTA from the test set of FinTabNet. The images above exhibit the ground truth, and the ones below shows the detection result.}
    \vspace{-4mm}
    \label{fig:error}
\end{figure}

As show in Figure~\ref{fig:error}, the major error types are summarized as follows: a) When two adjacent tables have identical formats, they may be detected as one because their combination also meets the criteria of aligning text in rows and columns; b) When there are standalone short sentences above the table, the model may mistakenly consider them as titles; c) When a table contains long text content, it could be incorrectly divided because the model misinterprets these long sentences as paragraphs between tables.
 These cases indicate that although learning text arrangement can effectively improve the model's robustness, it is still insufficient to fully address the diversity of tables. Combining multiple modalities of information, such as textual content and visual cues, may help handle such specific scenarios.

\section{Conclusion}
In this work, we focus on improving the stability of table detection models in the 0-shot setting when dealing with tables of different styles. We propose a light-weight and robust table detection method called TDeLTA, which learns text arrangement, classifies text blocks, and generates table boundaries. Experimental results show that TDeLTA, with only 3.1M parameters, achieves performance comparable to the state-of-the-arts, and shows the best robustness when faced with the cross-domain data under the 0-shot setting.


\bibliography{aaai24}

\end{document}